\documentclass[letterpaper, 10 pt, conference]{ieeeconf}  

\IEEEoverridecommandlockouts                              

\usepackage{cite}
\usepackage{amsmath,amssymb,amsfonts}
\usepackage{algorithmic}
\usepackage{graphicx}
\usepackage{textcomp}
\usepackage{xcolor}
\usepackage{amsmath}
\usepackage{graphicx}
\usepackage{footmisc}
\usepackage[bookmarks=false]{hyperref}
\usepackage{float}
\usepackage{authblk}
\usepackage{gensymb}
\usepackage{placeins}
\usepackage[ruled,linesnumbered]{algorithm2e} 
\usepackage{booktabs} 
\usepackage{balance} 
\usepackage{graphicx}
\usepackage{url}
\usepackage{caption}
\usepackage{epstopdf}
\usepackage{amssymb} 
\usepackage{bm}
\usepackage{pdfpages}
\usepackage{capt-of}
\usepackage{tabularx}
\usepackage{epsfig}
\usepackage{xcolor}
\usepackage{multirow}
\usepackage{nomencl}
\usepackage{amsmath}
\usepackage{siunitx}
\usepackage{subcaption}

\title{\LARGE \bf
3-D Motion Capture of an Unmodified Drone with \\Single-chip Millimeter Wave Radar
}

\author[$\dag$]{Peijun Zhao}
\author[$\S$]{Chris Xiaoxuan Lu}
\author[$\dag$]{Bing Wang}
\author[$\dag$]{Niki Trigoni}
\author[$\dag$]{Andrew Markham} 

\affil[$\dag$]{Department of Computer Science, University of Oxford, United Kingdom}
\affil[$\S$]{School of Informatics, University of Edinburgh, United Kingdom}

\begin{document}

\maketitle
\thispagestyle{empty}
\pagestyle{empty}

\begin{abstract}
Accurate motion capture of aerial robots in 3-D is a key enabler for autonomous operation in indoor environments such as warehouses or factories, as well as driving forward research in these areas. The most commonly used solutions at present are optical motion capture (e.g. VICON) and Ultrawideband (UWB), but these are costly and cumbersome to deploy, due to their requirement of multiple cameras/sensors spaced around the tracking area. They also require the drone to be modified to carry an active or passive marker. In this work, we present an inexpensive system that can be rapidly installed, based on single-chip millimeter wave (mmWave) radar. Importantly, the drone does not need to be modified or equipped with any markers, as we exploit the Doppler signals from the rotating propellers. Furthermore, 3-D tracking is possible from a single point, greatly simplifying deployment. We develop a novel deep neural network and demonstrate decimeter level 3-D tracking at 10Hz, achieving better performance than classical baselines. Our hope is that this low-cost system will act to catalyse inexpensive drone research and increased autonomy.

\end{abstract}

\section{Introduction}


Motion capture of  dynamic aerial robots (e.g. a small UAV or quadcopter drone) is a key enabling capability for na
vigation, path planning and autonomy. Motion capture in this context means the estimation of precise 3-D position in real-time.  In outdoor areas with a clear sky view, the use of RTK GPS, often combined with secondary sensors such as IMU, can provide excellent absolute positioning accuracy. 

Indoors or in GPS denied areas such as tunnels, alternative tracking approaches are typically employed. Broadly, these can be divided into optical or RF based techniques. VICON~\cite{al2015integration} is one of the most widely used optical motion capture platform. These systems provide high precision 3-D tracking of a reflective marker placed on a drone, provided that there is a good view from two or more spatially distributed cameras. These systems however are expensive (\$10k upwards), time-consuming to set up and calibrate, and require modifying the drone to be able to track it. To prevent dead-zones, a large number of cameras (e.g. 10) need to be placed around the tracking volume and typically are mounted at height (e.g. 2-3 m) to provide a wide field of view. A large amount of research has considered the use of UWB ranging as a multi-lateration approach\cite{guo2016ultra,macoir2019uwb}. These systems provide reasonable accuracy (10cm), but require the drone to carry a UWB transceiver to achieve pair-wise distance estimation. They also need a network of UWB transceivers distributed around the convex hull of the area to be tracked, leading to a time-consuming setup and calibration phase.

\begin{figure}[h]
    \centering
    \includegraphics[width=0.48\textwidth]{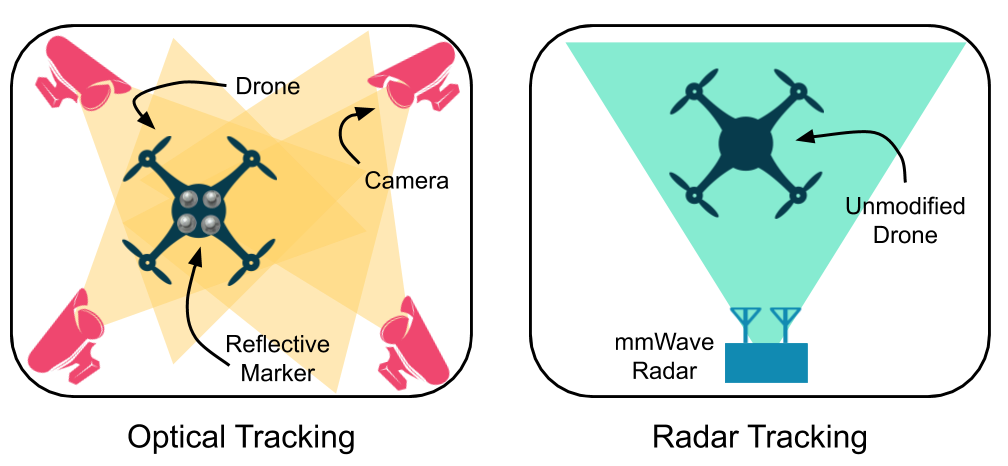}
    \caption{Compared with the optical tracking method, our proposed  system does not require any marker modification on the drone and only needs a low-cost mmWave radar.}
    \label{fig:open}
    \vspace{-20pt}
\end{figure}

In this paper we present a new approach for drone motion capture that is able to operate from a single point (i.e. it does not require multiple anchors to be carefully arranged around the tracking volume). As a result, it is easy to setup and deploy in new environments. mDrone is able to track an unmodified drone in 3-D with precision comparable to UWB ($<10cm$). Based on a single chip mmWave integrated radar transceiver, mDrone is low-cost ($<$\$300) and low-power, and has the capability of operating in the dark and other visually challenging conditions.


Our primary innovation is the use of single chip mmWave radar which brings its own set of unique challenges. Although radar has a long history of being used to track aeroplanes at extremely long ranges (hundreds of kms), the majority of solutions are military/aviation grade, large, costly and power-hungry. Recent advances in microchip fabrication have led to the availability of multi-channel, wideband radar transceivers that have been successfully used in diverse applications such as SLAM\cite{lu2020see,lu2020milliego,yassin2018mosaic,aladsani2019leveraging}, industrial automation\cite{jiang2020mmvib}, and human-computer interactions\cite{lien2016soli,zhao2019mid,singh2019radhar,meng2020gait,zhao2020heart}. However, a number of challenges arise from the problem of drone tracking. Firstly, the drone is unmodified, i.e. it does not carry any bright target e.g. a retroreflector. This makes it challenging to detect, but we demonstrate how to exploit measurements of the propeller's doppler velocity to refine detection. Secondly, the drone is physically large (e.g. 60 cm across), so accurately tracking its centre is difficult as the drone appears as a non-uniform blob in the radar returns. Thirdly, existing techniques are computationally expensive or inaccurate, precluding the ability for real-time motion capture. To address this, we introduce mDrone, a new deep learning pipeline and demonstrate its performance and robustness in comparison to five conventional signal processing baselines. We believe that this low-cost motion capture platform will not only be useful as a low-cost research and development tool, but also enable a number of downstream applications such as precision maintenance and inspection, robotic interaction, and increased autonomy in warehouses.

Our main contributions include:

\begin{itemize}
   \item We propose a deep learning method for estimation of drone position based on mmWave radar 
    \item We collected 100 sequences of data and evaluated our proposed algorithm against multiple conventional signal processing algorithms. The dataset will be released to the community together with the code. 
    \item Our system achieves a mean error of less than 10 cm in 3-D whilst operating at 10 Hz.
\end{itemize}

\section{Related Work}
\label{related_works}

\begin{table}[]
\vspace{10pt}
    \centering
    \begin{tabular}{c|c|c|c|c}
        \textbf{Solution} & \textbf{Cost~(\$)} & \textbf{Error~(cm)} & \textbf{Markers} & \textbf{Anchors}\\
        \hline
        Optical & 10000 & 1 & Yes & $\geq$2\\
        UWB & 3000 & 10 & Yes & $\geq$4\\
        mmWave Radar & 300 & 10 & No & 1\\
    \end{tabular}
    \caption{Comparison of Tracking Technologies.}
    \label{tab:solution_compare}
    \vspace{-15pt}
\end{table}

\subsection{Cooperative Localization}In cooperative tracking, modifications are made to the drone in order to track it. To achieve this, either markers or transceivers are installed on the drone itself. One of the most accurate 6-DOF cooperative tracking is based on Optical Tracking systems (e.g. Vicon). Hemispherical markers are attached to the body of the drone to form a rigid and spatially unique pattern, which are modeled and tracked by the system \cite{al2015integration, donald2018indoor}. Optical tracking systems are expensive (tens of thousands of dollars to equip a room with an optical tracking system) and time consuming to install and calibrate. Santos et. al. propose an optical tracking replacement system, for estimation of 6-DOF drone pose, with data from multiple sensors, including RGB-D camera and on-board IMU, based on computer vision and sensor fusion approaches~\cite{santos2017indoor}. 

UWB localization systems based on RF ranging and multilateration are widely used for their lower cost, whilst still providing accurate positioning (errors typically around $10cm$ or higher for outdoor scenarios)\cite{Tiemann2015,masiero2019comparison,guo2016ultra}. There are also many works that fuse UWB with other sensors for a better localization accuracy, such as UWB/IMU/Vision \cite{benini2013imu}, UWB/IMU~\cite{You2020}, and UWB/Radar~\cite{Zahran2018}. UWB techniques are also used for multi-UAV localization~\cite{Tiemann2017,Shule2020,Qi2020}.


The relative merits of our proposed mmWave Radar tracking technique compared with optical tracking and UWB tracking are summarized in TABLE.~\ref{tab:solution_compare}.



\subsection{Non-cooperative Localization}
Despite the fact that computer vision algorithms for object recognition and tracking with cameras are developing rapidly~\cite{8078539}, cameras do not work robustly under extreme light conditions, or in featureless areas. Researchers have been exploring other sensors for passive drone localization, including Light Detection And Ranging (LiDAR)~\cite{mirceta2018drone} and LAser Detection And Ranging (LADAR)~\cite{kim2018v}. 
Audio sensor arrays can be used for drone localization~\cite{chang2018surveillance}, which are based on TDOA principles, but they are sensitive to ambient noise and have limited range~\cite{8255737}.

Different types of radars are also widely used for drone detection and localization~\cite{guvencc2017detection, caris2017detection, wei2008detection}, amongst which, FMCW radars have attracted a great deal of research interest~\cite{drozdowicz201635}. In our work, we used a commercial-off-the-shelf mmWave FMCW radar, and focus on exploring the combination of signal processing and deep learning techniques to greatly improve short-range drone localization accuracy.




\section{Conventional Drone Tracking Pipeline}
\label{conventional}

\subsection{FMCW Radar Background}
\label{subsec:background}

\begin{figure}[h]
    \centering
    \includegraphics[width=0.38\textwidth]{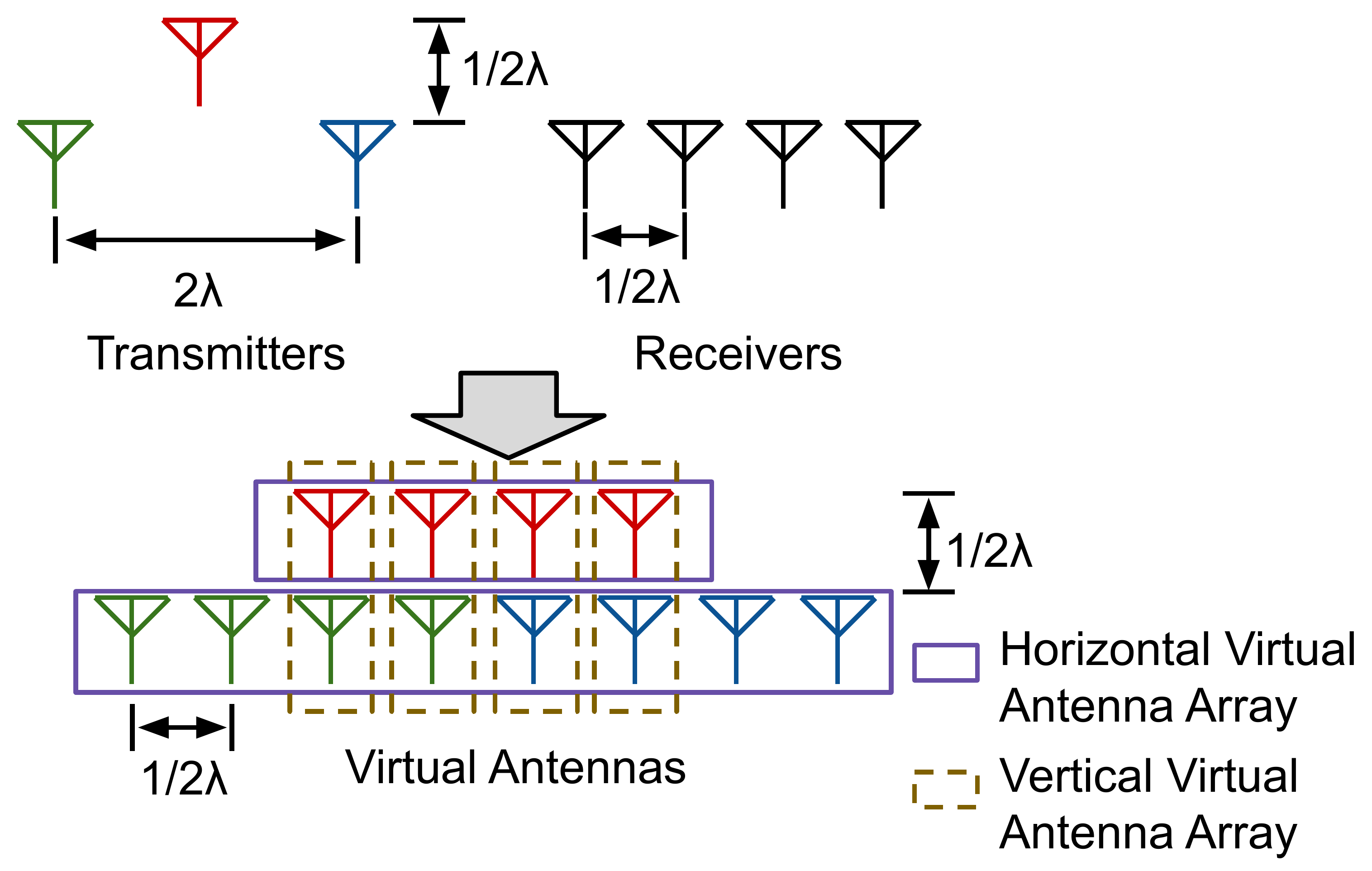}
    \caption{Texas Instruments IWR6843 mmWave Radar MIMO Virtual Antenna Array}
    \label{fig:virtual_antenna}
    \vspace{-10pt}
\end{figure}

We first introduce the principles and limitations behind conventional tracking pipelines. We consider a Frequency-Modulated Continuous-Wave (FMCW) Radar, which detects objects with electromagnetic wave chirps whose carrier frequency increases linearly with time. To demodulate, the reflected signal is mixed with the transmitting signal and produces an intermediate frequency (IF) signal, which is the difference between the transmitting and receiving signal. As the transmitting frequency is linearly increasing, the frequency shift of the IF signal is thus proportional to the distance of the object. The amplitude of the signal varies as a function of strength of the reflectivity. The distance to an object can be calculated by extracting peak frequency components  of the IF signal. This is typically accomplished using a signal processing technique such as a Fast Fourier Transform (FFT). As the radar emits chirps continuously, the phase shift between chirps at a particular frequency component of the IF signal can be used to estimate the speed of the object, caused by a Doppler shift. 

In this work, we consider the use of a Texas Instruments(TI) IWR6843 single chip radar, but the concepts presented here are generalizable to other platforms.
The TI IWR6843 radar features a 3Tx/4Rx MIMO antenna array, which forms 12 virtual antennas (VA) based on the combination of Tx-Rx pairs. The layout of the VA array is shown in Fig.~\ref{fig:virtual_antenna}. The phase differences between the antennas of the linear receiver arrays (horizontal and vertical) can be used to estimate the azimuth and elevation angle of arrival of different objects in the scene. For each frame, the radar is able to generate a data cube of complex numbers, with the axes respective to chirp index, samples, and virtual antennas.

\subsection{Sensitivity to Moving Objects}

FMCW radars are particularly good at detecting moving objects, because a small movement of the target leads to a large phase shift of the receiving signals. A UAV's rotating propellers can easily be detected and separated from the background static clutter. This is performed by a clutter removal algorithm that removes stationary objects from the FFT, as stationary objects have no phase shift. This property of the FMCW radar makes it uniquely suitable for tracking UAVs as we will demonstrate. 


\subsection{Tracking the Drone with a Radar Point Cloud}

\begin{figure}[h]
    \centering
    \includegraphics[width=0.5\textwidth]{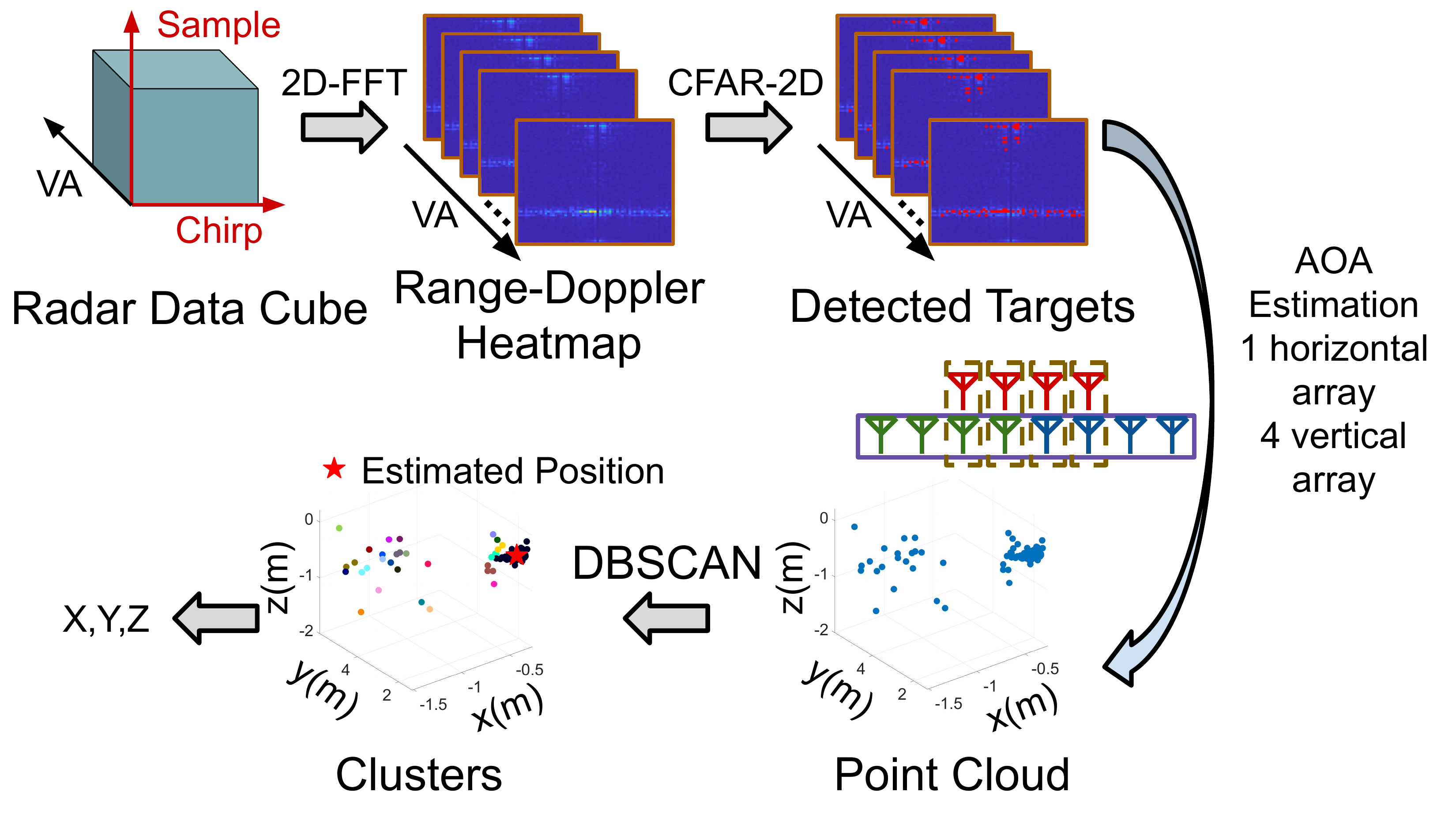}
    \caption{Radar Point Cloud pipeline}
    \label{fig:pt_pipeline}
    \vspace{-10pt}
\end{figure}

The mmWave radar is able to generate a 3D point cloud of the scene based on the principles detailed above, which can be used to localize the drone. As input, the radar data cube is first processed along the chirp-sample dimension for 2-D object detection, by forming a range-Doppler heatmap. The 2D-CFAR algorithm is used to extract dominant points. The phase differences between the virtual antennas corresponding to a detected point are then used to estimate the 3-D bearing angle of each point. These are then combined to produce a sparse 3D point cloud. The full pipeline is shown in Fig.~\ref{fig:pt_pipeline}. Using the 3D point cloud, a clustering algorithm such as DBScan is used to segment the point cloud into clusters. As the background clutter has been removed, the largest cluster corresponds to the drone, under the assumption that the scene is relatively stationary. By taking the centroid of the largest cluster, we can estimate the position of the drone.

\subsection{Tracking the Drone with Distance and Angle Estimation}

\begin{figure}[h]
\vspace{10pt}
    \centering
    \includegraphics[width=0.48\textwidth]{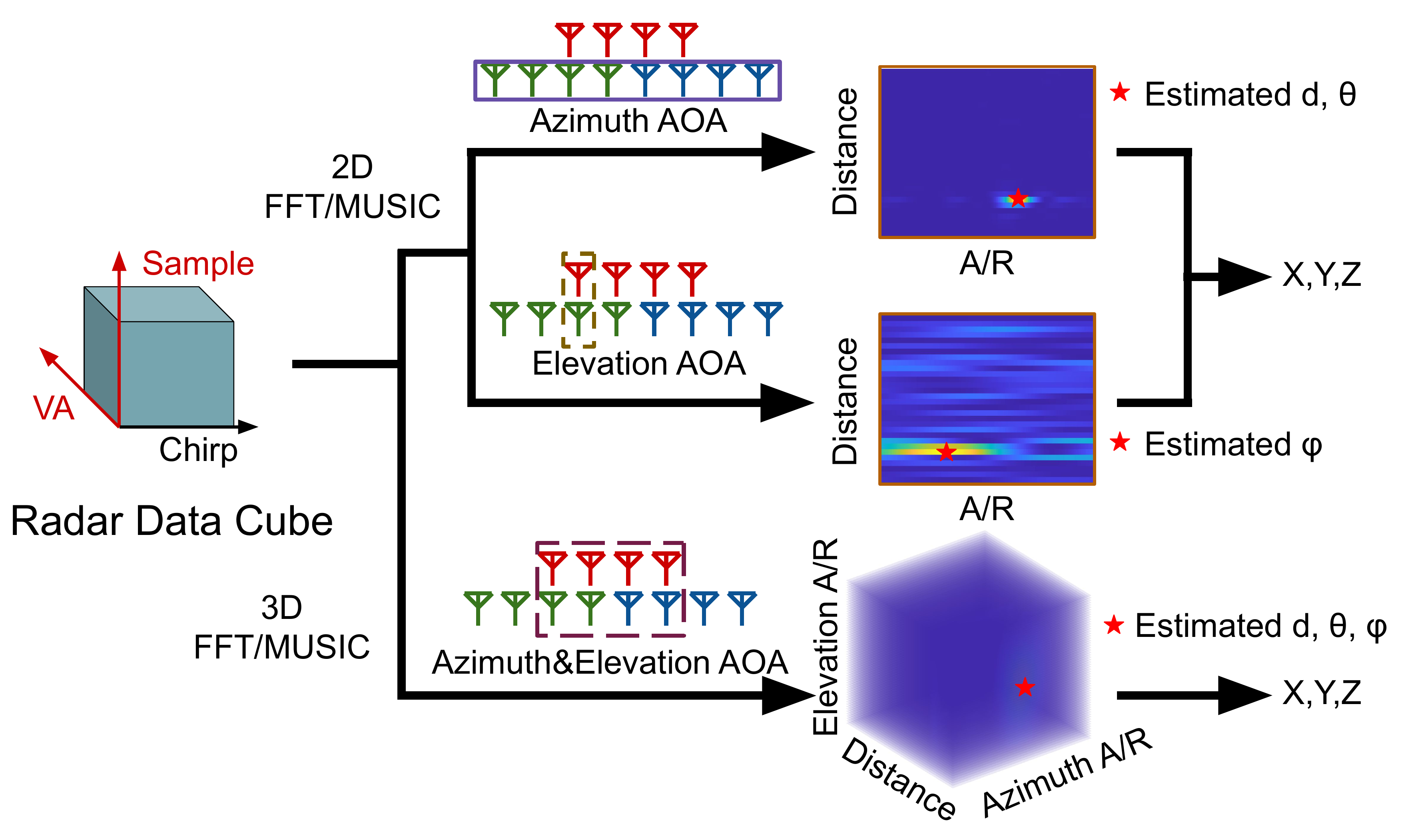}
    \caption{2D (top) and 3D (bottom) FFT and MUSIC pipelines.}
    \label{fig:fft_music}
    \vspace{-15pt}
\end{figure}

An alternative approach to estimate the location of the UAV can be performed by omitting the chirp axis of the radar cube, and only considering the Sample-VA plane. Conventional signal processing algorithms, such as the Fast Fourier Transform, or the super-resolution based MUSIC (MUltiple SIgnal Classification) algorithm\cite{schmidt1986multiple}, can be used to estimate the distance as well as both the azimuth and elevation angle of arrival (AoA) of the object. Given the distance and angle from multiple virtual antennas we are able to obtain the location of the drone. By repeating this process over the chirp axis of the radar cube, we can estimate the trajectory of the drone. 

As a refinement, we can perform 3D FFT or MUSIC directly on the cubic data formed by the Sample axis and the VA matrix shown in bottom pipeline of Fig.~\ref{fig:fft_music}, for a globally optimal estimation. The complexities of these algorithms are however exponential in the number of dimensions. 2D FFT/MUSIC based pipelines run significantly faster than 3D FFT/MUSIC based pipelines, but are more sensitive to noise and hence lack robustness and accuracy.





\subsection{Problems with Conventional Drone Tracking Pipeline}
Although tracking a drone based on conventional signal processing algorithms is  straightforward and easy to implement, these methods suffer from a number of problems which lead to low accuracy in practice. First of all, the drone is not a single point, but a complex object with multiple rotating points. Depending on the orientation of the drone relative to the antenna array, the signals from one or more propellors can be blocked or occluded by the drone body itself. This typically leads to a systematic localization error. In addition, there are many empirical parameters in conventional algorithms and it can be difficult to fine-tune and assign optimal values to those parameters. The localization results produced by conventional algorithms are not stable, and contain many outliers. This is because in some frames the noise or multipath scattering overwhelms the primary signal and is wrongly identified as the target. 

\section{Proposed Method}
\label{proposed_method}

\begin{figure*}[h]
\vspace{10pt}
    \centering
    \includegraphics[width=0.95\textwidth]{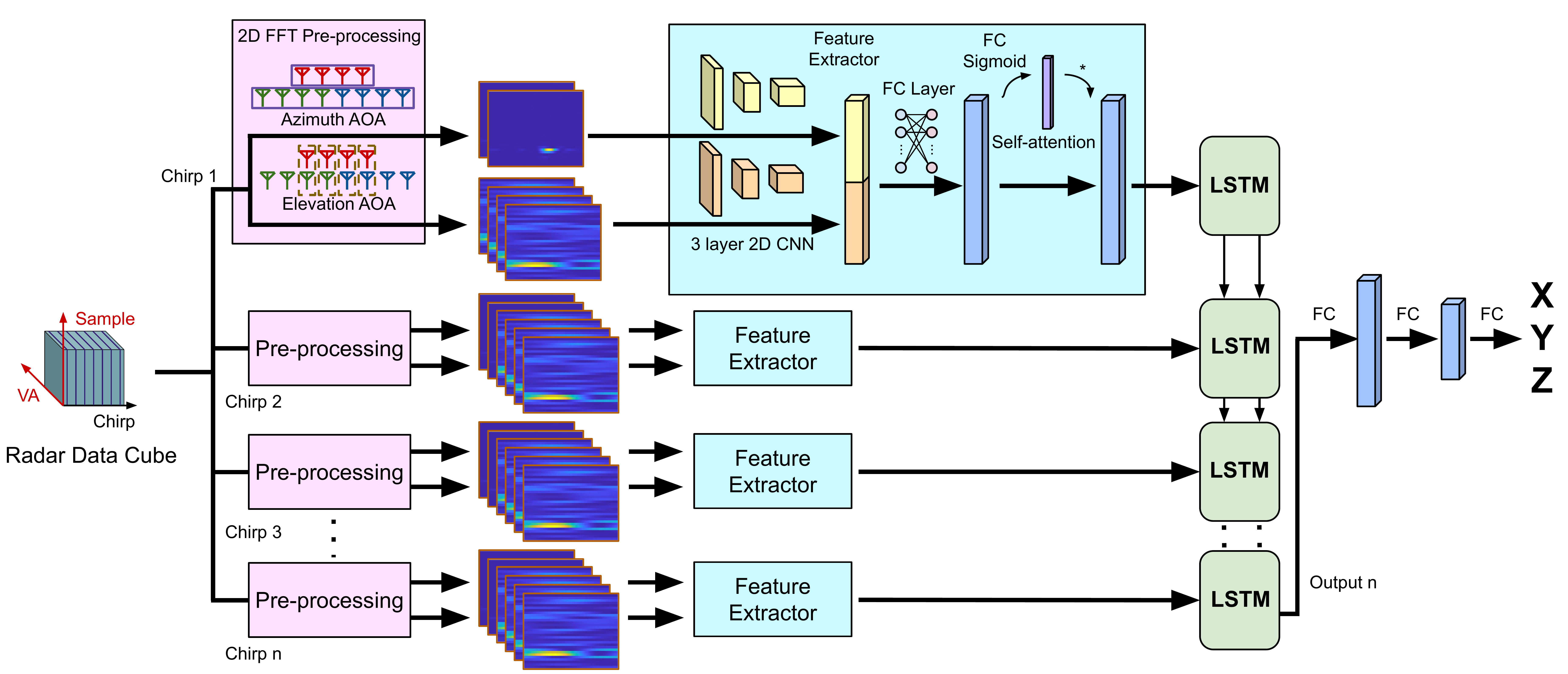}
    \caption{Proposed Method. For each chirp, the pre-processing module generates two range-azimuth heatmaps and four range-elevation heatmaps, which are sent into the Feature Extractor respectively. The feature vectors of multiple chirps in a frame are input into a LSTM network and the last output is used to estimate the 3D location.}
    \label{fig:proposed_pipeline}
    \vspace{-15pt}
\end{figure*}

Our method for drone localization contains two parts. Firstly, the radar cube data is preprocessed in range-elevation and range-azimuth heatmaps. Secondly, a deep neural network, consumes the heatmaps as input data and produces a 3D localization esimate. Fig.~\ref{fig:proposed_pipeline} shows the pipeline of the proposed method and the two parts are discussed below.

\subsection{Data Preprocessing}
The data processing part is similar to a conventional 2D FFT pipeline, except for that it produces 6 heatmaps in total (2 azimuth heatmaps and 4 elevation heatmaps), rather than just one azimuth heatmap and one elevation heatmap for target localization as in the conventional 2D FFT pipeline. In addition, heatmaps of multiple chirps with the data cube are extracted and sent into the neural network.

\subsection{Network Structure}


The neural network we propose contains two parts. The first part is a feature extractor, which takes range-azimuth heatmaps and range-elevation heatmaps as input. Two 3-layer 2D CNNs are used to extract an azimuth feature vector and an elevation feature vector. The two feature vectors are concatenated together and fed into a fully-connected layer to form an overall chirp feature vector which combines both azimuth and elevation information. A self-attention mechanism\cite{vaswani2017attention} is applied to the chirp feature, to assist in rejecting background noise by focussing on relevant features. The feature vectors of each chirp are sent into a LSTM network to provide an element of temporal smoothing. Three fully connected layers are then used to estimate the final output from the hidden layer of LSTM at the last timestamp.

\section{Implementation}
\label{sec_implementation}
\subsection{Data Collection}
\subsubsection{Experiment Setup}

\begin{figure}[h]
    \centering
    \includegraphics[width=0.3\textwidth]{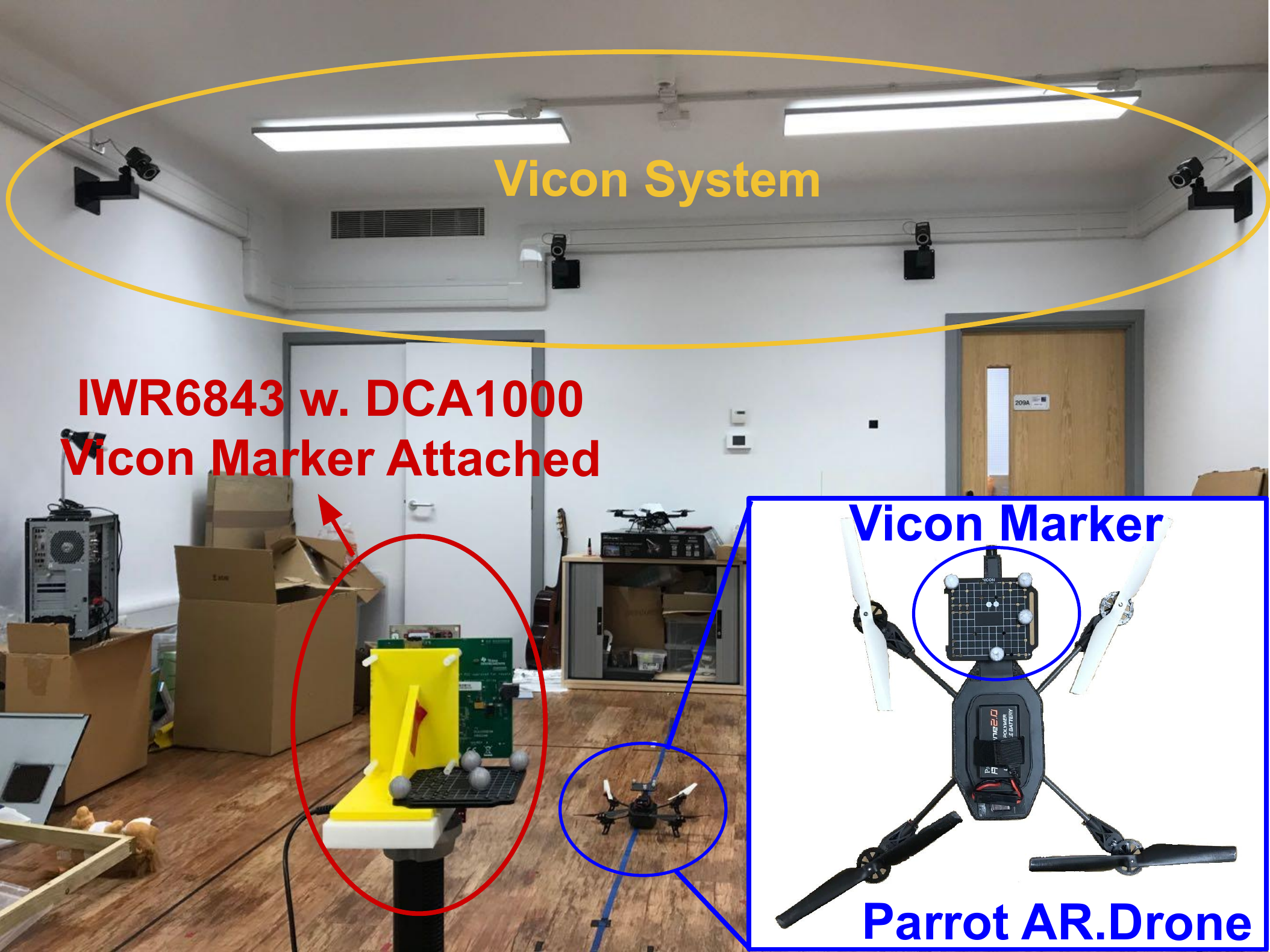}
    \caption{Experiment Settings}
    \label{fig:experiment_settings}
\end{figure}

We collected 100 1-minute sequences of a drone randomly flying in a room equipped with a Vicon system to provide accurate ground truth, as shown in Fig.~\ref{fig:experiment_settings}. For each sequence, the data collection begins after the drone is launched into the air so the 100 sequences purely contain the data of the drone flying.
We use a TI IWR6843 together with DCA1000EVM for raw data streaming. A desktop computer with Windows operating system is used for configuring the device and recording raw radar data, with mmWave Studio software. A laptop with Ubuntu operating system and ROS is used for Vicon data recording, and the Vicon data is later parsed to extract ground truth. 

\subsubsection{Coordinate System Definition}
The mmWave Radar is a right hand system, with y axis pointing forward, x axis pointing right, and z axis pointing upward.

\subsubsection{Time Synchronization}
Since the raw data streaming and Vicon data recording happen on two platforms respectively, we need to synchronize the timestamps of each frame in the mmWave data to the corresponding Vicon ground truth. The time of the two platforms are synchronized using NTP server. mmWave Studio is not able to provide timestamp logging for each frame, so we record the time when the radar starts and extrapolate the timestamp of each frame based on frame period. 


\subsection{Implementation}
We have 5 baselines in total, as introduced in Section.~\ref{conventional}, including Point Cloud based method, 2D FFT based method, 3D FFT based method, 2D MUSIC based method, and 3D MUSIC based method. All the conventional signal processing baselines are implemented in MATLAB. The size of the range FFT is 256, and the size of the Angle FFT is 180; in MUSIC, the range swept is 1m to 4m, with an interval of 0.1m, and the azimuth angle swept is 30\degree to 150\degree, with interval of 1\degree, and elevation angle -15\degree to 15\degree, with interval of 1\degree.

For our proposed method, the data preprocessing is done with MATLAB and the neural network is implemented in Python with PyTorch library.

\section{Evaluation}
\label{sec_evaluation}

\subsection{Localization Evaluation}
\begin{table}[]
\vspace{10pt}
    \centering
    \begin{tabular}{c|c|c|c|c}
        \textbf{Method} & \textbf{Avg} & \textbf{STD} & \textbf{Max} & \textbf{Min}\\
        \hline
        Point Cloud & 31.85 & 10.66 & 90.75 & 7.95\\
        2D FFT & 76.19 & 47.56 & 388.56 & 10.58\\
        3D FFT & 33.73 & 19.93 & 232.92 & 4.83\\
        2D MUSIC & 71.16 & 55.18 & 361.22 & 3.54\\
        3D MUSIC & 26.47 & 16.84 & 178.72 & 2.51\\
        \textbf{mDrone (Ours)} & \textbf{8.92} & \textbf{4.50} & \textbf{40.50} & \textbf{0.99}\\
    \end{tabular}
    \caption{Quantitative Localization Error Comparison (cm).}
    \label{tab:localization_result}
    \vspace{-15pt}
\end{table}

As our proposed methods involves deep learning techniques, we randomly picked 10 sequences as testing sequences, 10 sequences as validation sequences and the remaining 80 sequences as training sequences. All competing conventional methods are tested on the testing sequences. The deep neural network in our proposed pipeline is trained on the training sequences for 20 epochs and the parameters which produce the best validation result are saved and tested on the testing sequences. TABLE~\ref{tab:localization_result} summarizes the result of the localization evaluation.

From the result we can see that our proposed method significantly outperforms the competing signal processing methods. The average error of our proposed method is $\approx$3 times smaller the best competing baseline, which is 3D MUSIC. Besides lower average error, our proposed method is also much more robust than competing methods, for it has a Standard Deviation less than half of the Point Cloud based method, which is the most stable baseline. This proves that our proposed method can produce excellent results for UAV localization.

\begin{figure}[h]
    \centering
    \includegraphics[width=0.4\textwidth]{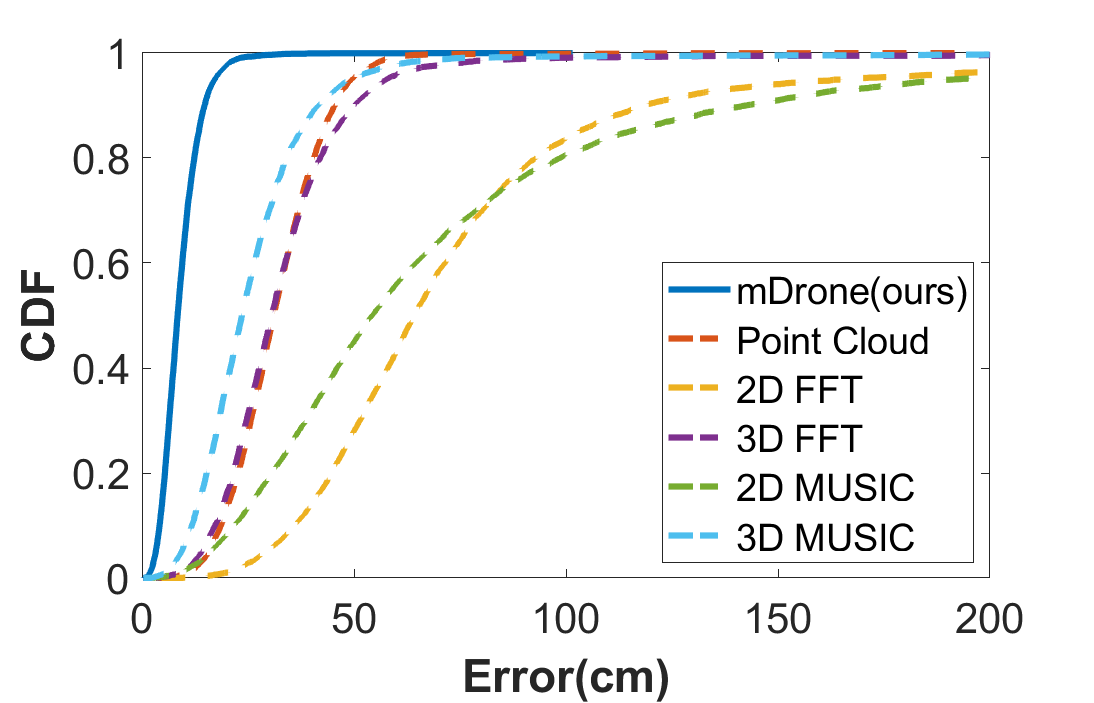}
    \caption{Cumulative Distribution Function (CDF) of the errors produced by different methods.}
    \label{fig:cdf}
\vspace{-10pt}
\end{figure}

To further investigate the localization error of different methods, we plot the Cumulative Distribution Function (CDF) of the errors of different methods, as shown in Fig.~\ref{fig:cdf}. The CDF shows that over 90\% of time, the error is  less than 15cm, which is significantly lower than the next best competitor (3D-MUSIC: 46cm).

\begin{figure}[t!] 
\begin{subfigure}{0.25\textwidth}
\includegraphics[width=\linewidth]{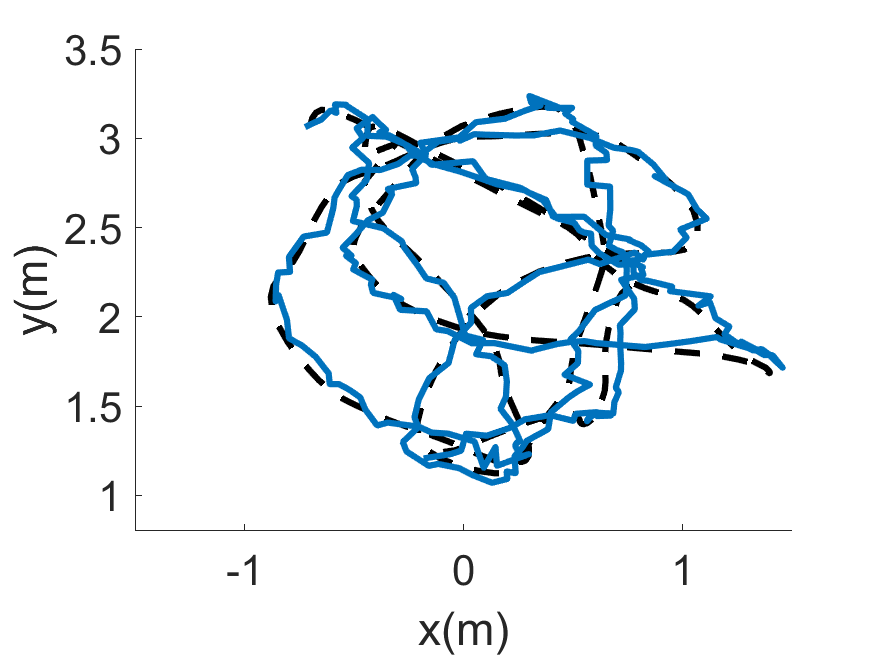}
\caption{mDrone (ours)} \label{fig:a}
\vspace{-5pt}
\end{subfigure}\hspace*{\fill}
\begin{subfigure}{0.25\textwidth}
\includegraphics[width=\linewidth]{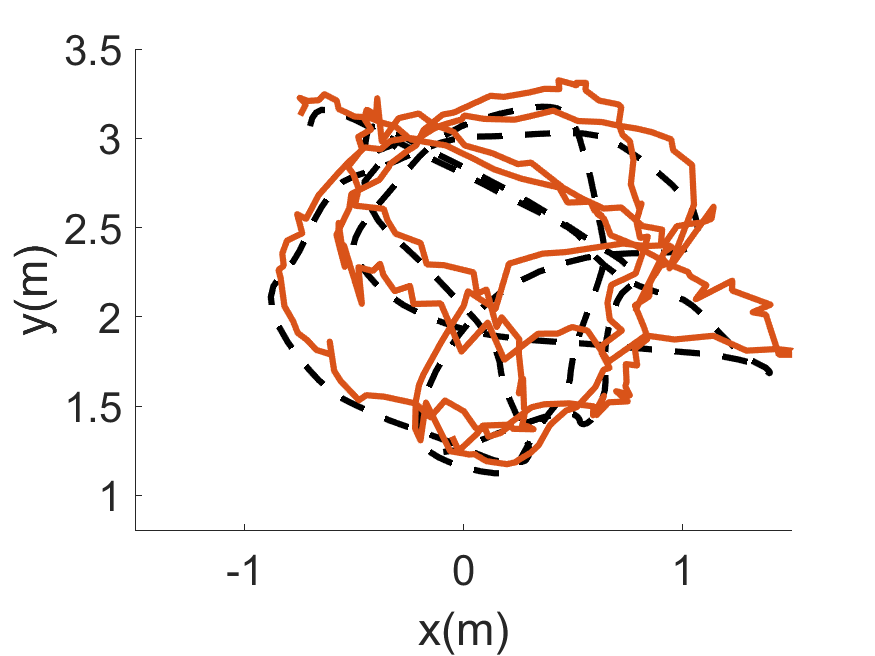}
\caption{Point Cloud} \label{fig:b}
\vspace{-5pt}
\end{subfigure}

\medskip
\begin{subfigure}{0.25\textwidth}
\includegraphics[width=\linewidth]{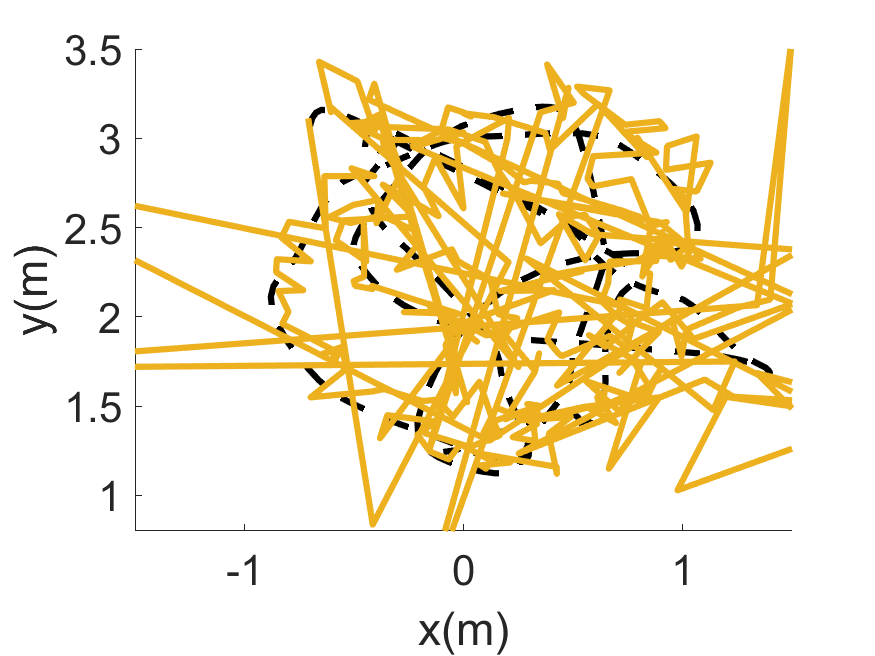}
\caption{2D FFT} \label{fig:c}
\vspace{-5pt}
\end{subfigure}\hspace*{\fill}
\begin{subfigure}{0.25\textwidth}
\includegraphics[width=\linewidth]{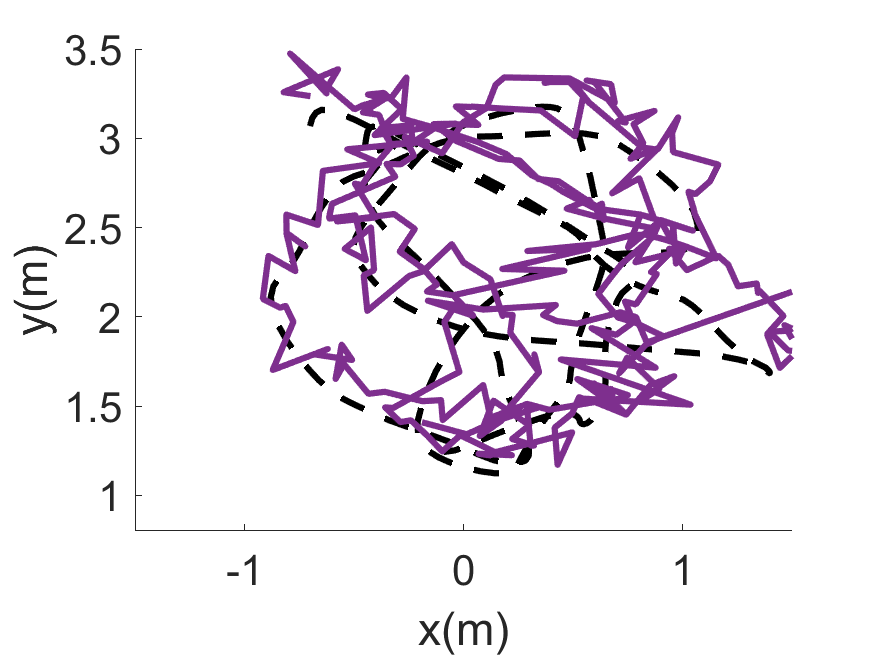}
\caption{3D FFT} \label{fig:d}
\vspace{-5pt}
\end{subfigure}

\medskip
\begin{subfigure}{0.25\textwidth}
\includegraphics[width=\linewidth]{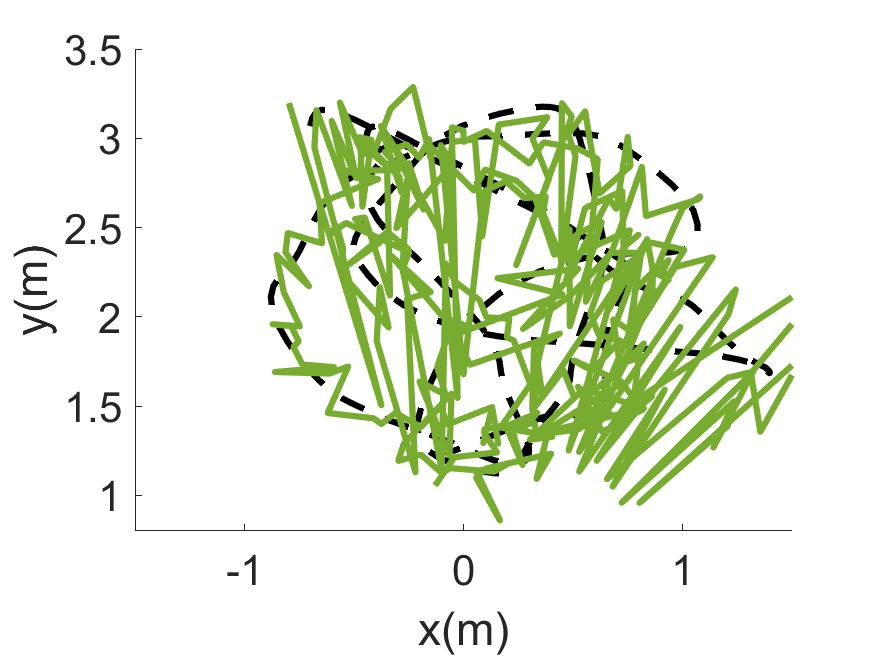}
\caption{2D MUSIC} \label{fig:e}
\vspace{-5pt}
\end{subfigure}\hspace*{\fill}
\begin{subfigure}{0.25\textwidth}
\includegraphics[width=\linewidth]{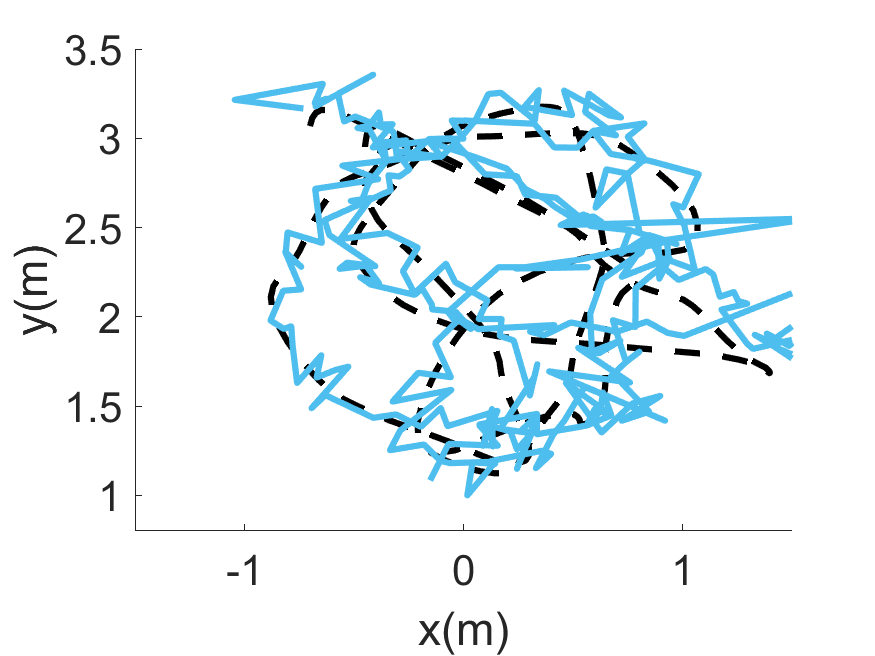}
\caption{3D MUSIC} \label{fig:f}
\vspace{-5pt}
\end{subfigure}

\caption{Qualitative localization comparison on a sample trajectory. Black dashed lines represent ground truth.} 
\label{fig:localization_sample}
\vspace{-10pt}
\end{figure}

A sample of the comparison of different techniques is shown in Fig.~\ref{fig:localization_sample}. 2D FFT and 2D MUSIC are not stable as they search for a single global maximum on the 2D heatmap, and often produce high errors. This is because a wrong prediction in either azimuth heatmap or elevation heatmap would result in a large 3-D overall error. 3D FFT and 3D MUSIC search for the peak in a 3D heatmap and hence more robust, with 3D MUSIC producing a better result thanks to the super-resolution feature. However, the localization accuracy is still worse than mDrone. The Point Cloud technique is more stable than 3D FFT and 3D MUSIC, but still significantly inferior compared to mDrone. Adding other techniques like sliding window average or Kalman Filter on the single radar cube localization results may further improve the stability and accuracy of mDrone, but it is beyond the topic of this paper and is not discussed here.

\subsection{Running Time Comparison}

\begin{table}[]
\vspace{10pt}
    \centering
    \begin{tabular}{c|c}
        \textbf{Method} & \textbf{Time per Frame (ms)}\\
        \hline
        Point Cloud & 101\\
        2D FFT & \textbf{5}\\
        3D FFT & 123\\
        2D MUSIC & 16477\\
        2D MUSIC (parallel)& 15426\\
        3D MUSIC & 282430\\
        3D MUSIC (parallel)& 220755\\
        \textbf{mDrone (Ours)} & 43\\
    \end{tabular}
    \caption{Running Time Comparison.}
    \label{tab:running_time}
\vspace{-15pt}
\end{table}

We compare the average running time of the conventional baselines as well as our proposed method. The MUSIC pipelines are implemented with both single thread and 16 thread parallel for speeding up. The experiment is performed with a AMD Ryzen 3800X CPU with 64GB main memory and a NVIDIA RTX2080Ti graphics card. The results are shown in TABLE.~\ref{tab:running_time}. mDrone is the second fastest among all the competing methods. The MUSIC algorithm based methods are shown to be extremely resource consuming as it takes more than 15 seconds for 2D music to process a radar data cube, and 220 seconds for 3D music, even in parallel mode. The Point cloud method and 3D FFT method take around ~100 ms for a radar data cube, and are more than 2 times slower than mDrone. The only method that is faster than ours, which is the 2D FFT method, produces much worse positioning results. mDrone takes around 43ms for a single frame on average. Note that the frame (radar data cube) period is 100ms in this work, which means that with proper implementation, mDrone is capable of running in real time at 10Hz. 

\subsection{Impact of Distance on Tracking Error}

\begin{figure}[h]
\vspace{-10pt}
    \centering
    \includegraphics[width=0.4\textwidth]{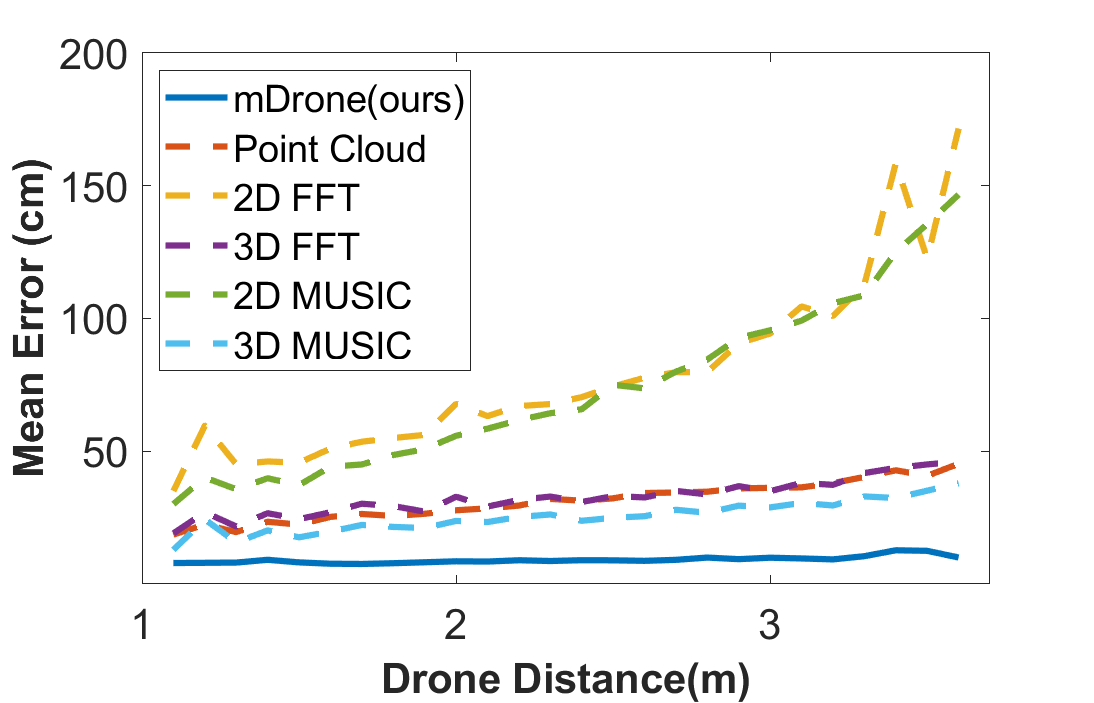}
    \caption{Mean Error v.s. Distance. Mean errors of conventional baselines increases with distance, while mean error of mDrone is relatively constant.}
    \label{fig:error_vs_distance}
    \vspace{-5pt}
\end{figure}

We calculated the mean error at different distances for each method, with the step of 0.1m. Results are shown in Fig.~\ref{fig:error_vs_distance}. The errors of conventional baseline methods, especially 2D FFT and 2D MUSIC, increases significantly with distance, as angular errors result in a larger overall position estimation error when the range is further. However, mDrone is able to achieve a relatively stable performance at different target distances, as the neural network takes the whole 2D range-angle heatmaps as inputs, rather than peak points, which could produce more reliable predictions.

\subsection{Ablation Study: Impact of Network Structure}

\begin{figure}[h]
    \centering
    \includegraphics[width=0.36\textwidth]{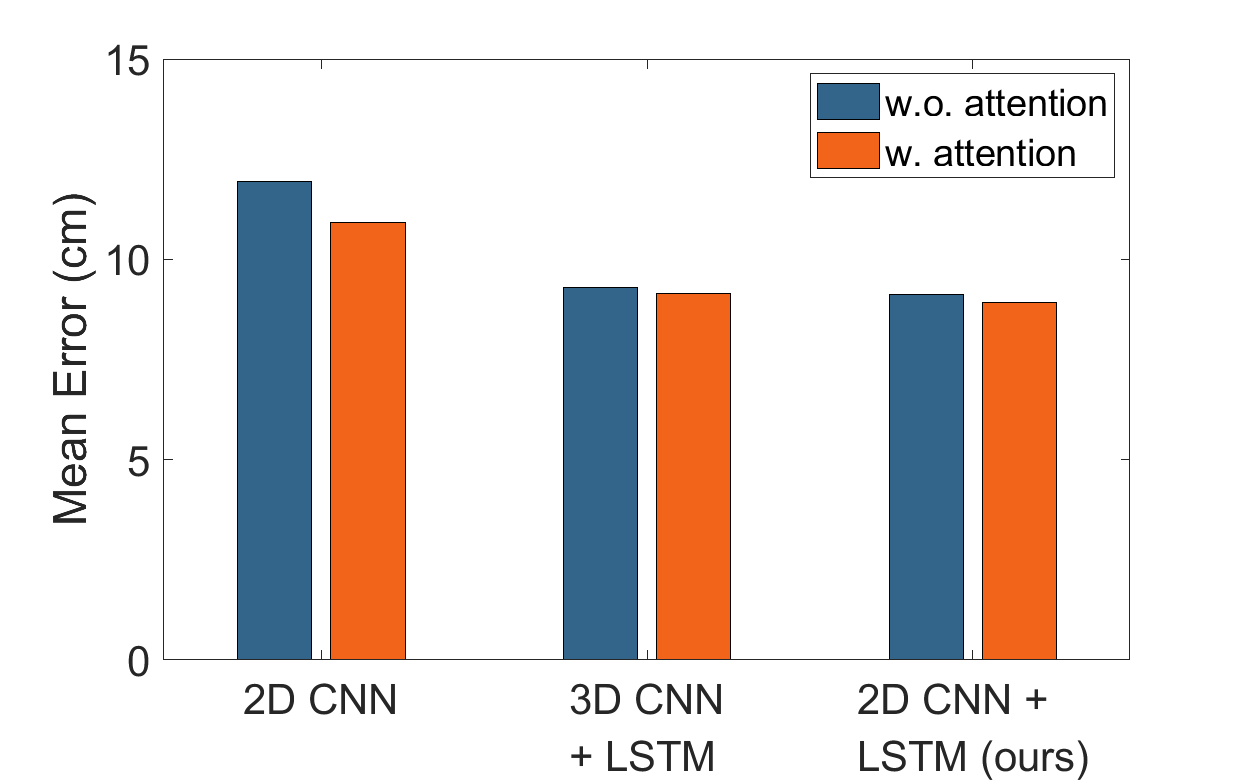}
    \caption{Comparison between Different Network Structures.}
    \label{fig:ablation_structure}
\vspace{-10pt}
\end{figure}

To show the functionality of different parts in our proposed neural network, we do the following three experiments:
\begin{itemize}
    \item To prove 2D CNN is a good choice for feature extraction, we replace the 2D CNN feature extractor with a 3D CNN feature extractor. (3D CNN + LSTM in Fig.~\ref{fig:ablation_structure})
    \item To show the effect of LSTM, we remove the LSTM and only use the features extracted by the feature extractor from the first chirp to estimate the localization result. (2D CNN in Fig.~\ref{fig:ablation_structure})
    \item To prove the effect of self-attention, we remove the self-attention module in each model. (Blue bars in Fig.~\ref{fig:ablation_structure})
\end{itemize}

Each of the above-mentioned models is trained with 80 training sequences for 20 epochs. The accuracy comparison is shown in Fig.~\ref{fig:ablation_structure}. As we can see from the result, the neural network produces better results when using a LSTM layer after the feature extractor than directly using FC layers to produce the final output by around 20\%. Using either 3D CNN or 2D CNN as feature extractor would produces similar results, while a 2D CNN is slightly better than 3D CNN by around 2.5\%. For all three structures, adding self-attention on the extracted feature increases the model performance by around 1.5\% to 8\%. The results as a whole though demonstrate the utility of data-driven techniques to track the moving drone, being relatively insensitive to the precise network structure.



\section{Discussion and Limitations}
\label{sec_discussion}

As mDrone is able to process one frame in less than 100ms, it could be used in real time in a 10Hz system with proper implementation. The localization performance of mDrone is relatively stable at different target distances, which implies that the tracking distance could be potentially extended further. 
We will be focusing on improving tracking range and the ability to generalize to completely different drone models (e.g., helicopter, hexacopter, mini-drones, etc.), as well as simultaneously tracking multiple drones in our future work.
\section{Conclusion}
\label{sec_conclustion}
In this work, we propose mDrone, a method of non-cooperative drone localization with a grounded millimeter wave radar, as an alternative for optical tracking and UWB. mmWave radar is very suitable for localizing drones for it is very sensitive to moving objects so the drone can be easily detected by it's propellers even when hovering. mDrone consists of a two-step tracking pipeline and is compared to several conventional signal processing baselines. Results show that our method could achieve a localization error of 8.92cm, which is more than 3 times better compared to the best baseline method. mDrone is also more stable than conventional signal processing methods, and is efficient enough to be used in real time. 










\bibliographystyle{IEEEtran}
\bibliography{ref.bib}

\end{document}